\documentclass{article}

     \usepackage[final,nonatbib]{neurips_2020}

\usepackage[utf8]{inputenc} 
\usepackage[T1]{fontenc}    
\usepackage{xr-hyper}
\usepackage{hyperref}       
\usepackage{url}            
\usepackage{booktabs}       
\usepackage{amsfonts}       
\usepackage{nicefrac}       
\usepackage{microtype}      

\usepackage{graphicx}
\usepackage{amsmath}
\usepackage{algorithm}
\usepackage{algorithmic}
\usepackage{wrapfig}
\usepackage{multirow}

\usepackage{xcolor}
\usepackage{subcaption}
\usepackage{amssymb}
\usepackage{bm}
\usepackage{bbm}
\usepackage{appendix}
\usepackage{enumitem}

\newcommand{\x}{\bm{x}}
\newcommand{\y}{\bm{y}}
\newcommand{\p}{\bm{p}}
\DeclareMathOperator{\argmax}{argmax}

\title{Self-Adaptive Training: beyond Empirical Risk Minimization}

\author{%
  Lang~Huang \\
  Peking University\\
  \texttt{laynehuang@pku.edu.cn} \\
  \And
  Chao Zhang\thanks{Corresponding authors} \\
  Peking University\\
  \texttt{c.zhang@pku.edu.cn} \\
  \And
  Hongyang Zhang$^{*}$ \\
  TTIC \\
  \texttt{hongyanz@ttic.edu} \\
}

\begin{document}

\maketitle

\begin{abstract}
We propose self-adaptive training---a new training algorithm that dynamically calibrates training process by model predictions without incurring extra computational cost---to improve generalization of deep learning for potentially corrupted training data. This problem is important to robustly learning from data that are corrupted by, e.g., random noises and adversarial examples. The standard empirical risk minimization (ERM) for such data, however, may easily overfit noises and thus suffers from sub-optimal performance. In this paper, we observe that model predictions can substantially benefit the training process: self-adaptive training significantly mitigates the overfitting issue and improves generalization over ERM under both random and adversarial noises. Besides, in sharp contrast to the recently-discovered double-descent phenomenon in ERM, self-adaptive training exhibits a single-descent error-capacity curve, indicating that such a phenomenon might be a result of overfitting of noises. Experiments on the CIFAR and ImageNet datasets verify the effectiveness of our approach in two applications: classification with label noise and selective classification. The code is available at \url{https://github.com/LayneH/self-adaptive-training}.

\end{abstract}

\section{Introduction}
Empirical Risk Minimization (ERM) has received significant attention due to its impressive generalization in various fields ~\cite{simonyan2014very,he2016deep}.
However, recent works~\cite{zhang2016understanding,NIPS2019_9336} cast doubt on the
traditional views on ERM: techniques such as uniform convergence might be unable to explain the generalization of deep neural networks, because ERM easily overfits training data even though the training data are partially or completely corrupted by random noises.

\begin{figure}[t]
    \centering
    \begin{subfigure}{\textwidth}
        \centering
        \includegraphics[width=\textwidth]{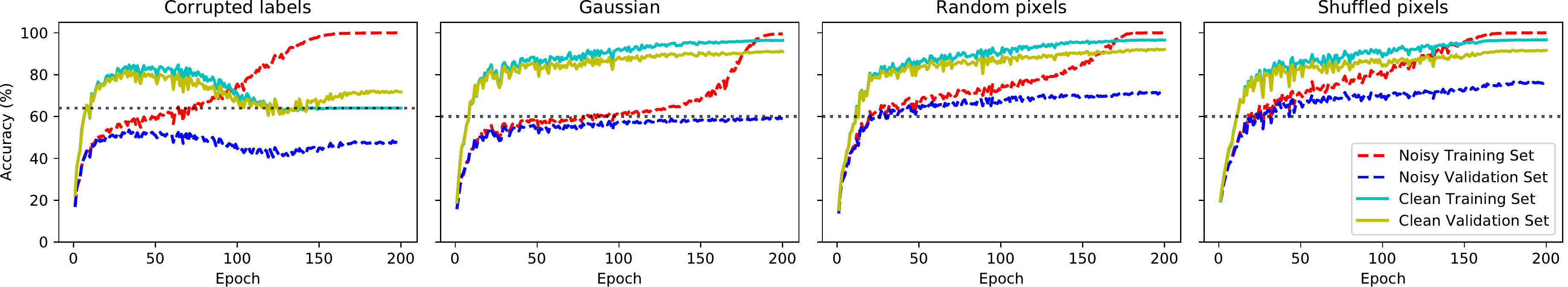}
        \vskip -0.0in
        \caption{Accuracy curves of model trained by ERM.}
        \label{fig:ce_acc_curve}
    \end{subfigure}
    \vskip 0.075 in
    \begin{subfigure}[]{\textwidth}
        \centering
        \includegraphics[width=\textwidth]{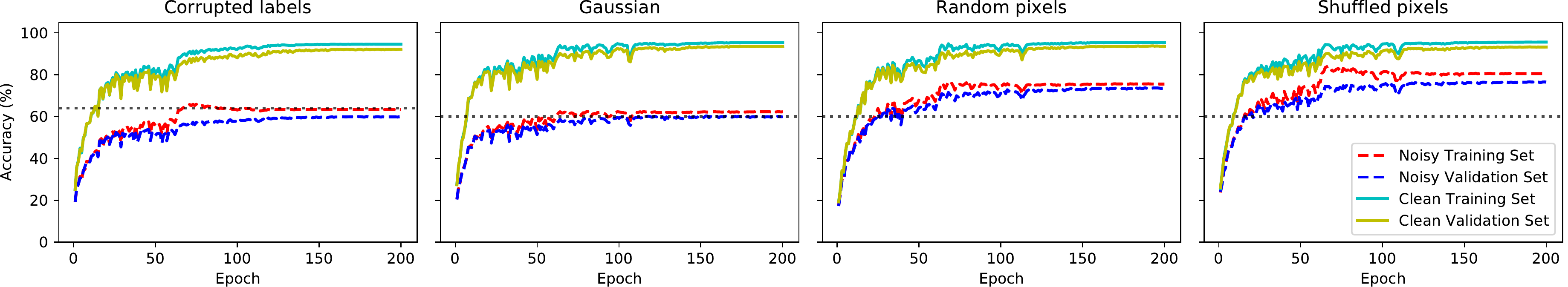}
        \vskip -0.0in
        \caption{Accuracy curves of model trained by our method.}
        \label{fig:wsc_acc_curve}
    \end{subfigure}
    \vskip -0.0in
    \caption{Accuracy curves of model trained on noisy CIFAR10 training set (corresponding to the red dashed curve) with 40\% corrupted data.
    The horizontal dotted line displays the percentage of clean data in the training sets.
    }
    \label{fig:acc_curve}
\end{figure}

To take a closer look at this phenomenon, we evaluate the generalization of ERM on the CIFAR10 dataset~\cite{krizhevsky2009cifar}
with 40\% of data being corrupted at random
(see Section~\ref{sec:approach} for details).
Figure~\ref{fig:ce_acc_curve} displays the accuracy curves
of ERM that are trained on the noisy training sets under four kinds of random corruptions: ERM easily overfits noisy training data and achieves nearly perfect training accuracy.
However, the four subfigures exhibit very different generalization behaviors which are indistinguishable by the accuracy curve on the noisy training set on its own.

Despite a large literature devoted to analyzing the phenomenon either in the theoretical or empirical manners, many fundamental questions remain unresolved.
To name a few, the work of~\cite{zhang2016understanding}
showed that early stopping can improve generalization.
On the theoretical front, the work of~\cite{li2019gradient} considered the label corruption setting, 
and proved that the first few training iterations fits the correct labels
and overfitting only occurs in the last few iterations: in Figure~\ref{fig:ce_acc_curve},
the accuracy increases in the early stage
and the generalization errors grow quickly after certain epochs.
Admittedly, stopping at early epoch improves generalization in the presence of label noises (see the first column in Figure \ref{fig:ce_acc_curve});
however, it remains unclear how to properly identify such an epoch. Moreover, the early-stop mechanism may significantly hurt the performance on the clean validation sets, as we can see in the second to the fourth columns of Figure~\ref{fig:ce_acc_curve}.

Our work is motivated by this fact and goes beyond ERM. We begin by making the following observations in the leftmost subfigure of Figure~\ref{fig:ce_acc_curve}:
the peak of accuracy curve on the clean training set (80\%) is much higher than 
the percentage of clean data in the noisy training set (60\%).
This finding was also previously reported
by~\cite{rolnick2017deep,guan2018said,li2019gradient} under label corruption and suggested that model predictions might be able to magnify useful underlying information in data.
We confirm this finding and show that the pattern occurs under various kinds of corruptions more broadly (see Figure~\ref{fig:ce_acc_curve}).
We thus propose \emph{self-adaptive training}, a carefully designed approach which dynamically uses model predictions as a guiding principle in the design of training algorithm.
Figure~\ref{fig:wsc_acc_curve} shows that our approach significantly alleviates the overfitting issue on the noisy training set, reduces the generalization error on the corrupted distributions, and improves the performance on the clean 
data. 

\subsection{Summary of our contributions}
\label{sec:contribution}
Our work sheds light on understanding generalization of deep neural networks under noise.

\begin{itemize}[leftmargin=*,itemsep=0.01em]
    \item
    We analyze the standard ERM training process of deep networks
    on four kinds of corruptions (see Figure~\ref{fig:ce_acc_curve}). We describe the failure scenarios of ERM and observe that useful information for classification has been distilled to model predictions in the first few epochs. This observation motivates us to propose self-adaptive training
    for robustly learning under noise.
    
    \item We show that self-adaptive training improves generalization under both label-wise and instance-wise random noises (see Figures~\ref{fig:acc_curve}~and~\ref{fig:gen_clean_errs}). Besides,
    self-adaptive training exhibits a single-descent error-capacity curve (see Figure~\ref{fig:double_descent}).
    This is in sharp contrast to the recently-discovered double-descent phenomenon in ERM which might be a result of overfitting of noise.
    
    \item While adversarial training may easily overfit adversarial noise, our approach mitigates the overfitting issue and improves adversarial accuracy by $\sim$3\% over the state-of-the-art (see Figure~\ref{fig:robust_acc}).
\end{itemize}

Our approach has two applications and advances the state-of-the-art by a significant gap.
\begin{itemize}[leftmargin=*,itemsep=0.01em]
    \item Classification with label noise, where the goal is to improve the performance of deep networks on clean test data in the presence of training label noise.
    On the CIFAR datasets, our approach achieves up to 9.3\% absolute improvement on the classification accuracy over the state-of-the-art.
    On the ImageNet dataset, our approach improves over ERM by 2\% under 40\% noise rate.
    
    \item Selective classification, which aims to trade prediction coverage off against classification accuracy.
    Our approach achieves up to 50\% relative improvement over the state-of-the-art on two datasets.
\end{itemize}

\noindent{\textbf{Differences between our methodology and existing works on robust learning}}\quad
Self-adaptive training consists of two components: a) the moving-average scheme that \emph{progressively} corrects problematic labels using model predictions; b) the re-weighting scheme that \emph{dynamically} puts less weights on the erroneous data. With the two components, our algorithm is robust to both instance-wise and label-wise noises, and is ready to combine with various training schemes such as natural and adversarial training, without incurring multiple rounds of training.
In contrast, a vast majority of works on learning from corrupted data follow a preprocessing-training fashion with an emphasis on the label-wise noise only: this line of research either discards samples based on disagreement between noisy labels and model predictions~\cite{brodley1996identifying,brodley1999identifying,zhu2003eliminating,nguyen2019self}, or corrects noisy labels~\cite{bagherinezhad2018label,tanaka2018joint}; \cite{teng1999correcting} investigated a more generic approach that corrects both label-wise and instance-wise noises. However, their approach inherently suffers from extra computational overhead.
Besides, unlike the general scheme in robust statistics~\cite{rousseeuw2005robust} and other re-weighting methods~\cite{jiang2018mentornet,ren2018learning} that use an additional optimization step to update the sample weights, our approach directly obtains the weights based on accumulated model predictions and thus is much more efficient.

\section{Improved Generalization of Deep Networks}
\label{sec:approach}

\subsection{Preliminary}
\label{sec:Corrupted data}
In this section, we conduct the experiments on the CIFAR10 dataset~\cite{krizhevsky2009cifar}, of which we split the original training data into a training set (consists of first 45,000 data pairs) and a validation set (consists of last 5,000 data pairs).
We consider four random noise schemes according to~\cite{zhang2016understanding}, where the data are \emph{partially} corrupted with probability $p$:
1)~\emph{Corrupted labels}. Labels are assigned uniformly at random;
2)~\emph{Gaussian}. Images are replaced by random Gaussian samples
with the same mean and standard deviation as the original image distribution;
3)~\emph{Random pixels}. Pixels of each image are shuffled using independent random permutations;
4)~\emph{Shuffled pixels}. Pixels of each image are shuffled using a fixed permutation pattern. We consider the performance on both the noisy and the clean sets (i.e., the original uncorrupted data), while the models can only have access to the noisy training sets.

\noindent\textbf{Notations}\quad
We consider $c$-class classification problem and denote the images by $\x_i\in \mathbb{R}^d$, labels by $\y_i\in \{0,1\}^c, \y_i^\intercal \mathbf{1}=1$.
The images $\x_i$ or labels $\y_i$ might be corrupted by one of the four schemes we have described. We denote the logits of the classifier (e.g., parameterized by a deep network) by $f(\cdot)$.

\subsection{Our approach: Self-Adaptive Training}

To alleviate the overfitting issue of ERM in Figure~\ref{fig:ce_acc_curve}, we present our approach to improve the generalization of deep networks on the corrupted data.

\noindent\textbf{The blessing of model predictions}\quad
As a warm-up algorithm, a straight-forward way to incorporate model predictions into the training process is to use a convex combination of labels and predictions as the training targets.
Concretely, given data pair $(\x_i, \y_i)$ and
prediction $\p_i = \mathrm{softmax}(f(\x_i))$,
we consider the training target $\bm{t}_i = \alpha\times \y_i + (1 - \alpha)\times \p_i$, where $\bm{t}_i\in [0,1]^c, \bm{t}_i^\intercal \mathbf{1}=1$.
We then minimize the cross entropy loss between $\p_i$ and $\bm{t}_i$ to update the classifier $f$ in each training iteration.
However, this naive algorithm suffers from multiple drawbacks:
1) model predictions are inaccurate in the early stage of training, and may be unstable in the presence of regularization such as data augmentation. This leads to instability of $\bm{t}_i$;
2) this scheme can assign at most $1 - \alpha$ weight on the true class when $\y_i$ is wrong. However, we aim to correct the erroneous labeling. In other words, we expect to assign nearly 100\% weight on the true class.

To overcome the drawbacks, we use the accumulated predictions to augment the training dynamics.
Formally, we initialize $\bm{t}_i \leftarrow \y_i$,
fix $\bm{t}_i$ in the first $\mathrm{E}_s$ training epochs, and update
$\bm{t}_i \leftarrow \alpha\times \bm{t}_{i} + (1 - \alpha)\times \p_i$
in each following training epoch.
The exponential-moving-average scheme alleviates the instability issue of model predictions, smooths out $\bm{t}_i$ during the training process and enables our algorithm to completely change the training labels if necessary.
Momentum term $\alpha$ controls the weight on the model predictions.
The number of initial epochs $\mathrm{E}_s$
allows the model to capture informative signals in the data set
and excludes ambiguous information that is
provided by model predictions in the early stage of training.

\noindent\textbf{Sample re-weighting}\quad
Based on the scheme presented above,
we introduce a simple yet effective sample re-weighting scheme on each sample.
Concretely, given training target $\bm{t}_i$, we set
$w_i = \max_{j} ~\bm{t}_{i,j}$.
The sample weight $w_i \in [\frac{1}{c}, 1]$
reveals the labeling confidence of this sample.
Intuitively, all samples are treated equally in the first $\mathrm{E}_s$ epochs.
As target $\bm{t}_i$ being updated, 
our algorithm pays less attention to potentially erroneous data
and learns more from potentially clean data.
This scheme also allows the corrupted samples to re-attain attention if they are confidently corrected.

\noindent\textbf{Putting everything together}\quad
We use stochastic gradient descent to minimize:
\begin{equation}
\label{eq:overall_loss}
    \mathcal{L}(f) = -\frac{1}{\sum_i w_i}\sum_i w_i \sum_j \bm{t}_{i,j}~\mathrm{log}~\p_{i,j}
\end{equation}
during the training process.
Here, the denominator normalizes per sample weights and stabilizes the loss scale. We name our approach \emph{Self-Adaptive Training} and display the pseudocode in Algorithm~\ref{alg:approach}. We fix the hyper-parameters $\mathrm{E}_s=60$, $\alpha=0.9$ by default if not specified. Our approach requires no modification to existing network architecture and incurs almost no extra computational cost.

\begin{algorithm}[h]
\small
\caption{Self-Adaptive Training}
\label{alg:approach}
\begin{algorithmic}[1]
  \REQUIRE Data $\{(\x_i, \y_i)\}_n$, initial targets $\{\bm{t}_{i}\}_n = \{\y_{i}\}_n$, batch size $m$, classifier $f$, $\mathrm{E}_s = 60$, $\alpha = 0.9$
    \REPEAT
      \STATE Fetch mini-batch data $\{(\x_i, \bm{t}_{i})\}_m$ at current epoch $e$
      \FOR{$i=1$ {\bfseries to} $m$  (in parallel)}
        \STATE $\p_i = \mathrm{softmax}(f(\x_i))$
        \STATE \textbf{if} $e > \mathrm{E}_s$ \textbf{then} $\bm{t}_{i} = \alpha\times \bm{t}_{i} + (1 - \alpha)\times \p_i$
        \STATE $w_{i} = \max_{j} ~\bm{t}_{i,j}$
      \ENDFOR
      \STATE Update $f$ by SGD on $\mathcal{L}(f) = -\frac{1}{\sum_i w_i}\sum_i w_i \sum_j \bm{t}_{i,j}~ \mathrm{log}~\p_{i,j}$
    \UNTIL{end of training}
\end{algorithmic}
\end{algorithm}

\begin{figure}
\centering
\includegraphics[width=\textwidth]{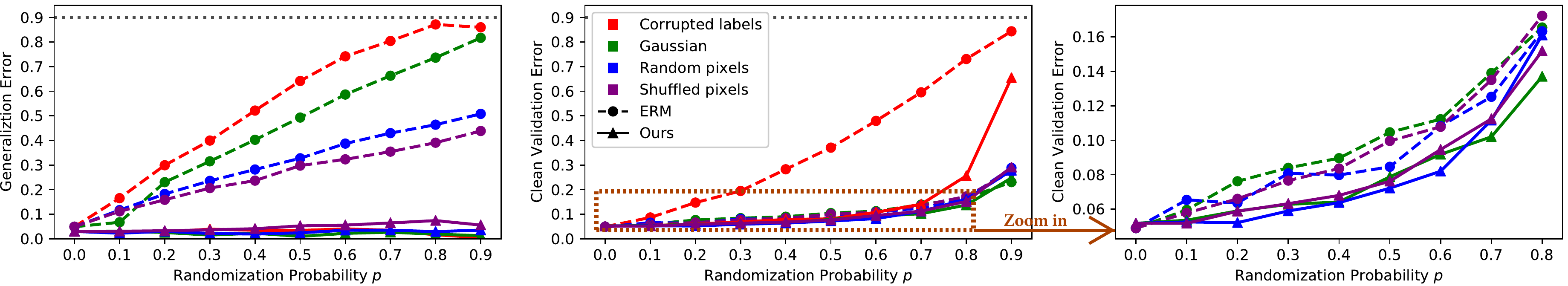}
\vskip -0.0in
\caption{
Generalization error and clean validation error under four random noises (represented by different colors) for ERM (the dashed curves) and our approach (the solid curves) on CIFAR10.
We zoom-in the dashed rectangle region and display it in the third column for clear demonstration.
}
\label{fig:gen_clean_errs}
\end{figure}

\subsection{Improved generalization of self-adaptive training under random noise}
We consider noise scheme (including noise type and noise level) and model capacity as two factors that affect the generalization of deep networks under random noise. We analyze self-adaptive training by varying one of the two factors while fixing the other.

\noindent\textbf{Varying noise schemes}\quad
We use ResNet-34~\cite{he2016deep} and rerun the same experiments in Figure~\ref{fig:ce_acc_curve} by replacing ERM with our approach. In Figure~\ref{fig:wsc_acc_curve}, we plot the accuracy curves of models trained with our approach on four corrupted training sets and compare with Figure~\ref{fig:ce_acc_curve}.
We highlight the following observations.

\begin{itemize}[leftmargin=*,itemsep=0.01em]
\item Our approach mitigates the overfitting issue in deep networks. The accuracy curves on noisy training sets (i.e., the red dashed curves in Figure~\ref{fig:wsc_acc_curve}) 
nearly converge to the percentage of clean data in the training sets, and do not reach perfect accuracy.

\item The generalization errors of self-adaptive training (the gap between the red and blue dashed curves in Figure~\ref{fig:wsc_acc_curve}) are much smaller than Figure~\ref{fig:ce_acc_curve}. We further confirm this observation by displaying the generalization errors of the models trained on the four noisy training sets under various noise rates in the leftmost subfigure of Figure~\ref{fig:gen_clean_errs}. Generalization errors of ERM consistently grow as we increase the injected noise level. In contrast, our approach significantly reduces the generalization errors across all noise levels from 0\% (no noise) to 90\% (overwhelming noise).

\item The accuracy on the clean sets (cyan and yellow solid curves in Figure~\ref{fig:wsc_acc_curve}) is monotonously increasing and converges to higher values than their correspondence in Figure~\ref{fig:ce_acc_curve}. We also show the clean validation errors in the right two subfigures in Figure~\ref{fig:gen_clean_errs}. The figures show that the error of self-adaptive training is consistently much smaller than that of ERM.
\end{itemize}

\noindent\textbf{Varying model capacity}\quad
We notice that such analysis is related to a recent-discovered intriguing phenomenon~\cite{belkin2018reconciling,nakkiran2019deep} in modern machine learning models: as the capacity of model increases, the test error initially decreases, then increases, and finally shows a second descent. This phenomenon is termed \emph{double descent} \cite{belkin2018reconciling} and has been widely observed in deep networks~\cite{nakkiran2019deep}. To evaluate the double-descent phenomenon on self-adaptive training, we follow exactly the same experimental settings as \cite{nakkiran2019deep}: we vary the width parameter of ResNet-18~\cite{he2016deep} and train the networks on the CIFAR10 dataset with 15\% training label being corrupted at random (details are given in Appendix~\ref{sec:setup_dd}).
Figure~\ref{fig:double_descent} shows the curves of test error. It shows that self-adaptive training overall achieves much lower test error than that of ERM in most cases. Besides, we observe that the curve of ERM clearly exhibits the double-descent phenomenon, while the curve of our approach is monotonously decreasing as the model capacity increases. Since the double-descent phenomenon may vanish when label noise is absent~\cite{nakkiran2019deep}, our experiment indicates that this phenomenon may be a result of overfitting of noises and we can bypass it by a proper design of training process such as the self-adaptive training.

\noindent\textbf{Potential failure scenarios}\quad
We notice that self-adaptive training could perform worse than ERM when using extremely small models that underfit the training data. Under such cases, the models do not have enough capacity to capture sufficient information, incorporating their ambiguous prediction may even hinder the training dynamics. However, as shown in Figure~\ref{fig:double_descent}, the ERM can only outperform our self-adaptive training in some extreme cases that the models are $10\times$ smaller than the standard ResNet-18, indicating that our method can work well in most realistic settings.

\begin{figure}[t]
\minipage{0.575\textwidth}
  \centering
  \includegraphics[width=.65\linewidth]{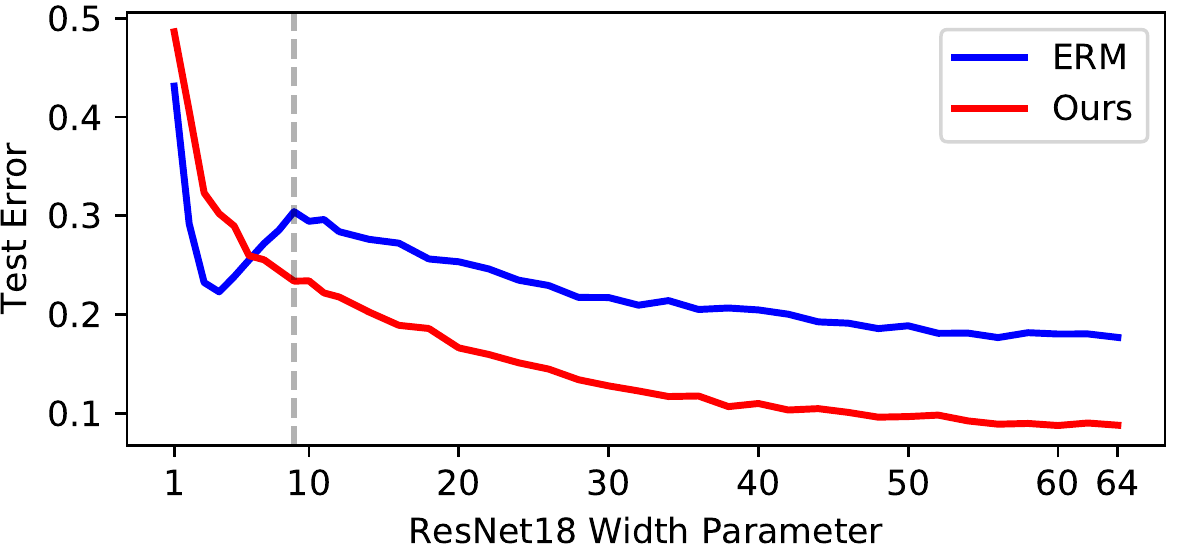}
  \caption{
    Double-descent ERM \emph{vs.} single-descent self-adaptive training on the error-capacity curve. 
    The model of width 64 corresponds to standard ResNet-18. The vertical dashed line represents the interpolation threshold~\cite{belkin2018reconciling,nakkiran2019deep}.
  }
  \label{fig:double_descent}
\endminipage
\hfill
\minipage{0.4\textwidth}
  \centering
  \includegraphics[width=.935\linewidth]{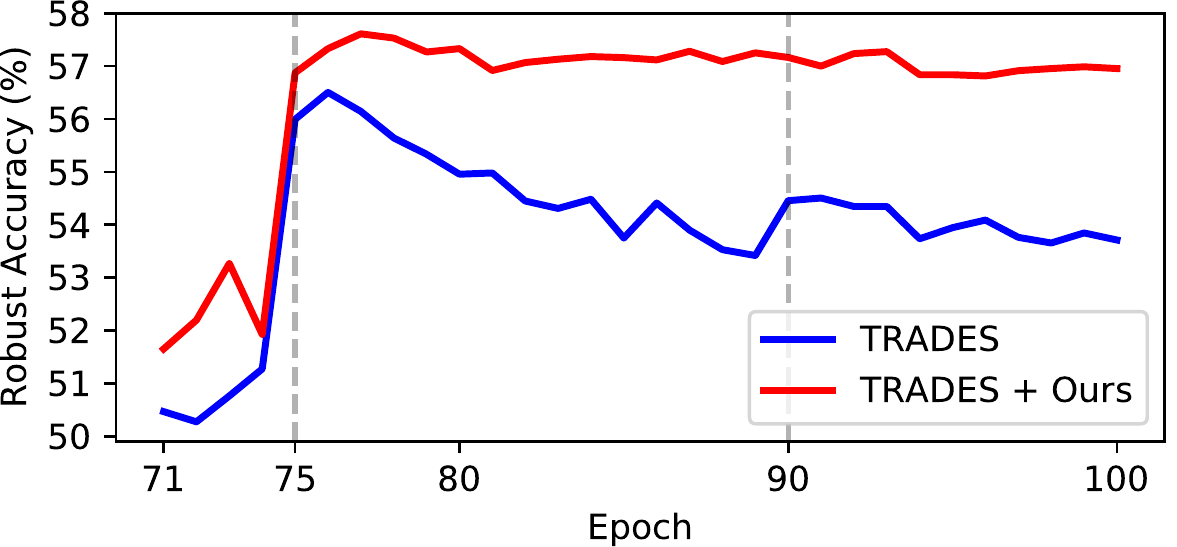}
  \caption{
    Robust Accuracy (\%) on CIFAR10 test set under white box $\ell_\infty$ PGD-20 attack ($\epsilon$=0.031). The vertical dashed lines indicate learning rate decay.
  }
  \label{fig:robust_acc}
\endminipage
\end{figure}

\subsection{Improved generalization of self-adaptive training under adversarial noise}
Adversarial noise~\cite{szegedy2013intriguing} is different from the random noise in that the noise is model-dependent and imperceptible to humans.
We use the state-of-the-art adversarial training algorithm TRADES \cite{zhang2019theoretically} as our baseline to evaluate the performance of self-adaptive training under adversarial noise.
Algorithmally, TRADES minimizes
\begin{equation}
\label{eq:trades}
\mathbb{E}_{\x,\y}\Bigg\{ \mathrm{CE}(\p(\x), \y) + \max_{\|\widetilde{\x}-\x\|_\infty\le\epsilon}\mathrm{KL}(\p(\x), \p(\widetilde{\x}))/\lambda\Bigg\},
\end{equation}
where $\p(\cdot)$ is the model prediction, $\epsilon$ is the maximal allowed perturbation, CE stands for the cross entropy, KL stands for the Kullback–Leibler divergence, and the hyper-parameter $\lambda$ controls the trade-off between robustness and accuracy.
We replace the CE term in TRADES loss with our method.
The models are evaluated using robust accuracy $\frac{1}{n}\sum_i \mathbbm{1}\{ \argmax~\p(\widetilde{\x}_i)=\argmax~\y_i \}$, where adversarial example $\widetilde{\x}$ are generated by white box $\ell_{\infty}$ projected gradient descent (PGD) attack \cite{madry2017towards} with $\epsilon$ = 0.031, perturbation steps of 20. We set the initial learning rate as 0.1 and decay it by a factor of 0.1 in epochs 75 and 90, respectively. 
We choose $1/\lambda=6.0$ as suggested by~\cite{zhang2019theoretically} and use $\mathrm{E}_s$ = 70, $\alpha$ = 0.9 for our approach.
Experimental details are given in Appendix~\ref{sec:setup_adv}.

We display the robust accuracy on CIFAR10 test set after $\mathrm{E}_s$ = 70 epochs in Figure~\ref{fig:robust_acc}. It shows that the robust accuracy of TRADES reaches its highest value around the epoch of first learning rate decay (epoch 75) and decreases later, which suggests that overfitting might happen if we train the model without early stopping.
On the other hand, self-adaptive training considerably mitigates the overfitting issue in the adversarial training and consistently improves the robust accuracy of TRADES by 1\%$\sim$3\%, which indicates that our method can improve the generalization in the presence of adversarial noise.
\section{Application I: Classification with Label Noise}
Given improved generalization of self-adaptive training over ERM under noise, we provide applications of our approach which outperforms the state-of-the-art with a significant gap.

\subsection{Problem formulation}
Given a set of noisy training data $\{(\x_i, \widetilde{\y}_i)\}_n \in \mathcal{\widetilde{D}}$, where $\mathcal{\widetilde{D}}$ is the distribution of noisy data and $\widetilde{\y}_i$ is the noisy label for each uncorrupted sample $\x_i$, the goal is to be robust to the label noises in the training data and improve the classification performance on clean test data that are sampled from clean distribution $\mathcal{D}$.

\subsection{Experiments on CIFAR datasets}
\label{sec:exp_label_noise}
\noindent\textbf{Setup}\quad
We consider the case that the labels are assigned uniformly at random with different noise rates.
Following prior works~\cite{zhang2018gce,thulasidasan2019dac}, we conduct the experiments on the CIFAR10 and CIFAR100 datasets~\cite{krizhevsky2009cifar} using ResNet-34~\cite{he2016deep} and Wide ResNet 28~\cite{zagoruyko2016wide} as our base classifiers.
The networks are implemented on PyTorch~\cite{paszke2019pytorch} and optimized using SGD with initial learning rate of 0.1, momentum of 0.9, weight decay of 0.0005, batch size of 256, total training epochs of 200. The learning rate is decayed to zero using cosine annealing schedule~\cite{loshchilov2016sgdr}. We use data augmentation of random horizontal flipping and cropping. We report the average performance over 3 trials.

\noindent\textbf{Main results}\quad
We summarize the experiments in Table~\ref{tab:noisy_cls_all}. Most of the results are cited from original papers  when they are under the same experiment settings; the results of Label Smoothing~\cite{szegedy2016rethinking},  Mixup~\cite{zhang2017mixup}, Joint Opt~\cite{tanaka2018joint} and SCE~\cite{wang2019symmetric} are reproduced by rerunning the official open-sourced implementations.
From the table, we can see that our approach outperforms the state-of-the-art methods in most entries by 1\% $\sim$ 9\% on both CIFAR10 and CIFAR100 datasets. Notably, unlike Joint Opt, DAC and SELF methods that require multiple iterations of training, our method enjoys the same computational budget as ERM.

\begin{table}[t!]
\caption{Test Accuracy (\%) on CIFAR datasets with various levels of uniform label noise injected to training set. We compare with previous works under exactly the same experiment settings.
It shows that in most settings, self-adaptive training improves over the state-of-the-art as significant as $9$\%.
}
\label{tab:noisy_cls_all}
\begin{center}
\begin{small}
\resizebox{\columnwidth}{!}{
\begin{tabular}{llcccccccc}
\toprule
\multirow{2}{*}{Backbone} &  & \multicolumn{4}{c}{CIFAR10} & \multicolumn{4}{c}{CIFAR100} \\
& Label Noise Rate  & 0.2 & 0.4 & 0.6 & 0.8 & 0.2 & 0.4 & 0.6 & 0.8   \\
\midrule
\multirow{10}{*}{ResNet-34} & ERM + Early Stopping & 85.57 & 81.82 & 76.43 & 60.99 & 63.70 & 48.60 & 37.86 & 17.28 \\
& Label Smoothing~\cite{szegedy2016rethinking}& 85.64 & 71.59 & 50.51 & 28.19 & 67.44 & 53.84 & 33.01 & 9.74 \\
& Forward $\hat{T}$~\cite{patrini2017making}  & 87.99 & 83.25 & 74.96 & 54.64 & 39.19 & 31.05 & 19.12 & 8.99 \\
& Mixup~\cite{zhang2017mixup}                 & 93.58 & 89.46 & 78.32 & 66.32 & 69.31 & 58.12 & 41.10 & 18.77 \\
& Trunc $\mathcal{L}_q$~\cite{zhang2018gce}   & 89.70 & 87.62 & 82.70 & 67.92 & 67.61 & 62.64 & 54.04 & 29.60 \\
& Joint Opt~\cite{tanaka2018joint}            & 92.25 & 90.79 & 86.87 & 69.16 & 58.15 & 54.81 & 47.94 & 17.18 \\
& SCE~\cite{wang2019symmetric}                 & 90.15 & 86.74 & 80.80 & 46.28 & 71.26 & 66.41 & 57.43 & 26.41 \\
& DAC~\cite{thulasidasan2019dac}              & 92.91 & 90.71 & 86.30 & 74.84 & 73.55 & 66.92 & 57.17 & 32.16 \\
& SELF~\cite{nguyen2019self}                  & - & 91.13 & - & 63.59 & - & 66.71 & - & 35.56 \\
& Ours    & \textbf{94.14} & \textbf{92.64} & \textbf{89.23} & \textbf{78.58} & \textbf{75.77} & \textbf{71.38} & \textbf{62.69} & \textbf{38.72} \\
\midrule
\multirow{5}{*}{WRN28-10} & ERM + Early Stopping & 87.86 & 83.40 & 76.92 & 63.54 & 68.46 & 55.43 & 40.78 & 20.25 \\
& MentorNet~\cite{jiang2018mentornet} & 92.0 & 89.0 & - & 49.0 & 73.0 & 68.0 & - & 35.0 \\
& DAC~\cite{thulasidasan2019dac} & 93.25 & 90.93 & 87.58 & 70.80 & 75.75 & 68.20 & 59.44 & 34.06 \\
& SELF~\cite{nguyen2019self}    & - & \textbf{93.34} & - & 67.41 & - & 72.48 & - & 42.06 \\
& Ours    & \textbf{94.84} & 93.23 & \textbf{89.42} & \textbf{80.13} & \textbf{77.71} & \textbf{72.60} & \textbf{64.87} & \textbf{44.17} \\
\bottomrule
\end{tabular}
}
\end{small}
\end{center}
\end{table}

\begin{table}[t]
\caption{Ablation study on CIFAR datasets in terms of classification Accuracy (\%).}
\label{tab:ablation}
\begin{subtable}{.48\textwidth}
\caption{Influence of the two components of our approach.}
\label{tab:ablation_ema_reweight}
\begin{center}
\begin{small}
\setlength{\tabcolsep}{1.5mm}{
\begin{tabular}{lcccc}
\toprule
 & \multicolumn{2}{c}{CIFAR10} & \multicolumn{2}{c}{CIFAR100}\\
Noise Rate              & 0.4   & 0.8   & 0.4   & 0.8   \\
\midrule
Ours                    & \textbf{92.64} & \textbf{78.58} & \textbf{71.38} & \textbf{38.72} \\
- Re-weighting          & 92.49 & 78.10 & 69.52 & 36.78 \\
- Moving Average        & 72.00 & 28.17 & 50.93 & 11.57 \\
\bottomrule
\end{tabular}
}
\end{small}
\end{center}
\end{subtable}
\vspace{-0.1in}
\hfill
\begin{subtable}{.48\textwidth}
\caption{Parameters sensitivity when label noise of 40\% is injected to CIFAR10 training set.}
\label{tab:ablation_alpha_es}
\begin{center}
\begin{small}
\setlength{\tabcolsep}{1.5mm}{
\begin{tabular}{lccccc}
\toprule
$\alpha$ & 0.6 & 0.8 & 0.9 & 0.95 & 0.99 \\
\hline
Fix $\mathrm{E}_s$=60 & 90.17 & 91.91 & \textbf{92.64} & 92.54 & 84.38 \\
\hline\hline
$\mathrm{E}_s$ & 20 & 40 & 60 & 80 & 100 \\
\hline
Fix $\alpha$=0.9 & 89.58 & 91.89 & \textbf{92.64} & 92.26 & 88.83 \\
\bottomrule
\end{tabular}
}
\end{small}
\end{center}
\end{subtable}
\end{table}

\noindent\textbf{Ablation study and parameter sensitivity}\quad
First, we report the performance of ERM equipped with simple early stopping scheme in the first row of Table~\ref{tab:noisy_cls_all}. We observe that our approach achieves substantial improvements over this baseline. This demonstrates that simply early stopping the training process is a sub-optimal solution.
Then, we further report the influences of two individual components of our approach: Exponential Moving Average (EMA) and sample re-weighting scheme. As displayed in Table~\ref{tab:ablation_ema_reweight}, removing any component considerably hurts the performance under all noise rates and removing EMA scheme leads to a significant performance drop. This suggests that properly incorporating model predictions is important in our approach.
Finally, we analyze the sensitivity of our approach to the parameters $\alpha$ and $\mathrm{E}_s$ in Table~\ref{tab:ablation_alpha_es} (and also Table~\ref{tab:ablation_alpha_es_supp} of Appendix). The performance is stable for various choices of $\alpha$ and $\mathrm{E}_s$, indicating that our approach is insensitive to the hyper-parameter tuning.

\subsection{Experiments on ImageNet dataset}
\begin{wraptable}{r}{0.275\textwidth}
\caption{Top1 Accuracy (\%) on ImageNet validation set.}
\label{tab:imagenet}
\begin{center}
\begin{small}
\begin{tabular}{lcc}
\toprule
Noise Rate  & 0.0 & 0.4 \\
\midrule
ERM     & 76.8 & 69.5  \\
Ours    & \textbf{77.2} & \textbf{71.5}  \\
\bottomrule
\end{tabular}
\end{small}
\end{center}
\end{wraptable}
The work of~\cite{russakovsky2015imagenet} suggested that ImageNet dataset \cite{deng2009imagenet} contains annotation errors even after several rounds of cleaning. Therefore, in this subsection, we use ResNet-50~\cite{he2016deep} to evaluate self-adaptive training on the ImageNet under both standard setup (i.e., using original labels) and the case that 40\% training labels are corrupted. We provide the experimental details in Appendix~\ref{sec:setup_imagenet} and report model performance on the ImageNet validation set in terms of top1 accuracy in Table~\ref{tab:imagenet}. We see that self-adaptive training consistently improves the ERM baseline by a considerable margin (e.g., 2\% when 40\% labels are corrupted), which validates the effectiveness of our approach on large-scale dataset.

\subsection{Further inspection on self-adaptive training}
\begin{figure}[t]
\minipage{0.475\textwidth}
  \centering
  \includegraphics[width=.9\textwidth]{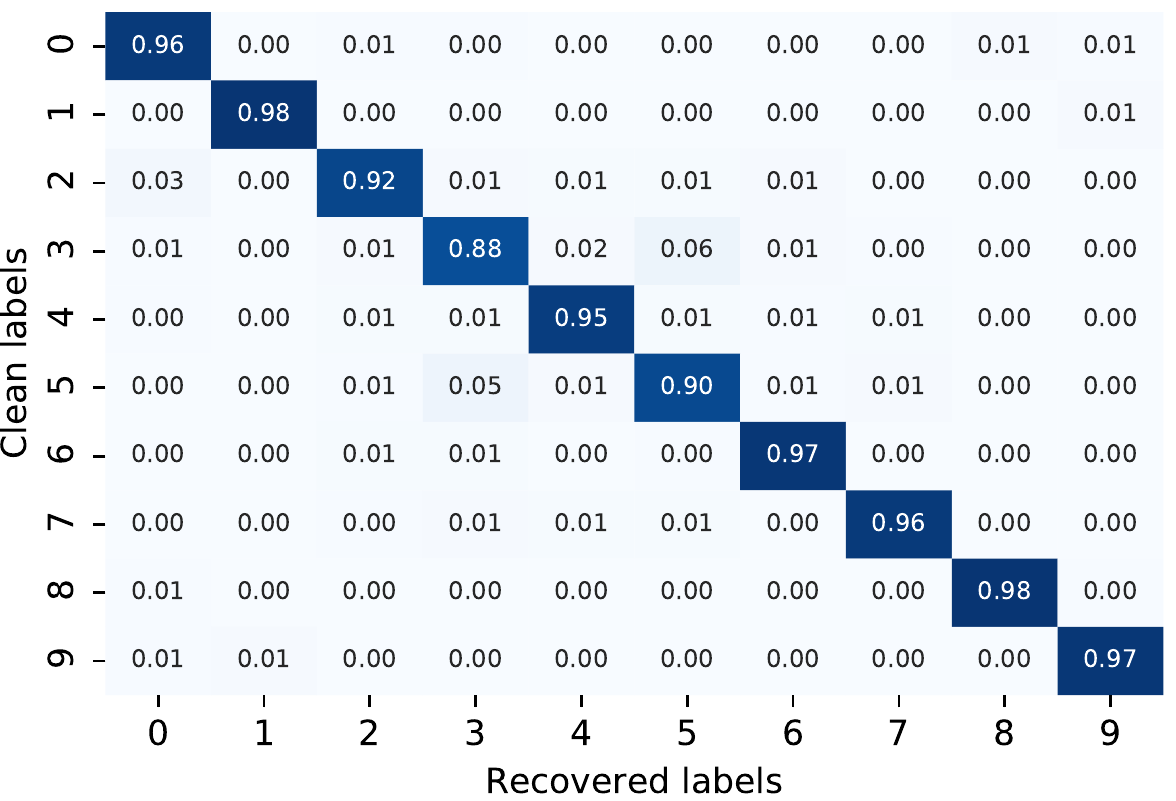}
  \caption{Confusion matrix of recovered labels w.r.t clean labels on CIFAR10 training set with 40\% of label noise.}
  \label{fig:conf_mat}
\endminipage
\hfill
\minipage{0.475\textwidth}
  \centering
  \includegraphics[width=.9\textwidth]{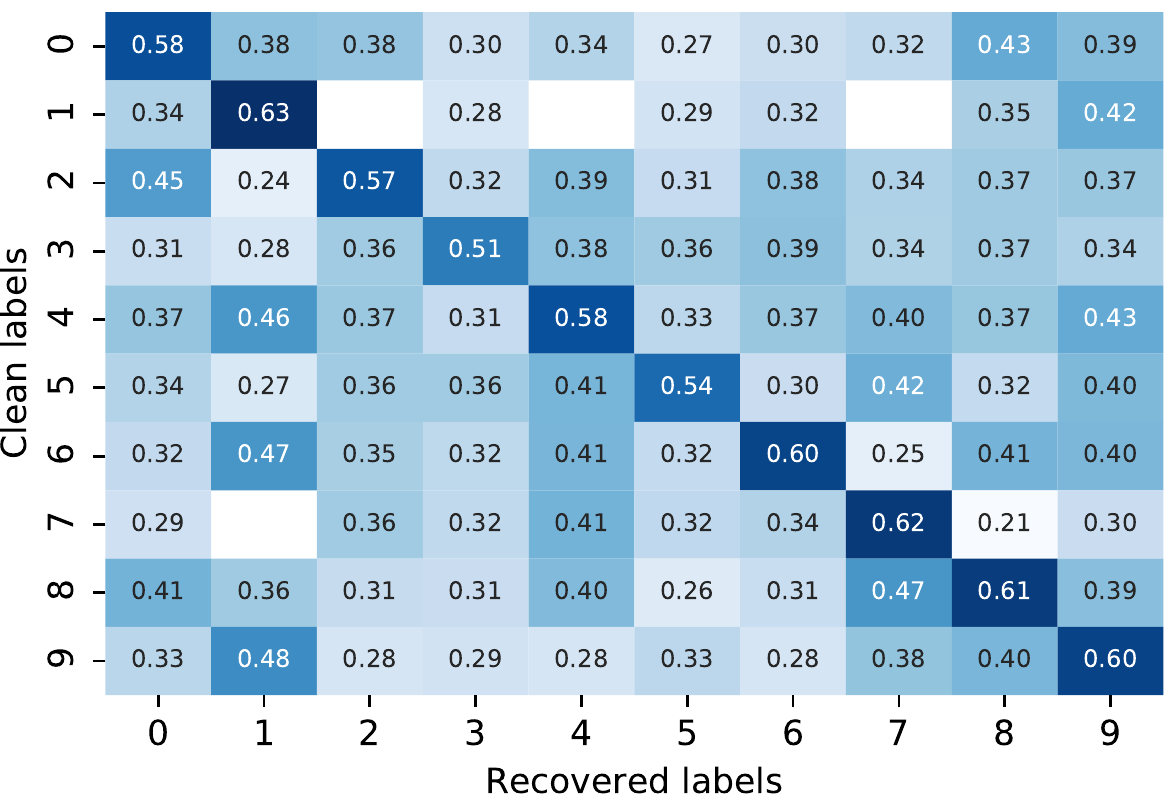}
  \caption{Average sample weights $w_i$ under various labels. The white areas indicate that no sample lies in the cell.}
  \label{fig:weights}
\endminipage
\end{figure}

\noindent\textbf{Label recovery}\quad
We demonstrate that our approach is able to recover the true labels from noisy training labels: we obtain the recovered labels by the moving average targets $\bm{t}_i$ and compute the recovered accuracy as $\frac{1}{n}\sum_i \mathbbm{1}\{ \argmax~\y_i = \argmax~\bm{t}_i \}$, where $\y_i$ is the clean label of each training sample. When 40\% label are corrupted in the CIFAR10 and ImageNet training set, our approach successfully corrects a huge amount of labels and obtains recovered accuracy of 94.6\% and 81.1\%, respectively. We also display the confusion matrix of recovered labels w.r.t the clean labels on CIFAR10 in Figure~\ref{fig:conf_mat}, from which see that our approach performs impressively well for all classes.

\noindent\textbf{Sample weights}\quad
Following the same procedure, we display the average sample weights in Figure~\ref{fig:weights}. In the figure, the $(i, j)$-th block contains the average weight of samples with clean label $i$ and recovered label $j$, the white areas represent the case that no sample lies in the cell. We see that the weights on the diagonal blocks are clearly higher than those on non-diagonal blocks. The figure indicates that, aside from impressive ability to recover the correct labels, self-adaptive training could properly down-weight the noisy examples.

\section{Application II: Selective Classification}
\subsection{Problem formulation}
Selective classification, a.k.a. classification with rejection,
trades classifier coverage off against accuracy~\cite{el2010foundations},
where the coverage is defined as the fraction of classified samples in the dataset; the classifier is allowed to output ``don't know'' for certain samples.
The task focuses on noise-free setting and
allows classifier to abstain on potential out-of-distribution samples 
or samples lies in the tail of data distribution, that is,
making prediction only on samples with confidence.
Formally, a selective classifier is a composition of two functions $(f, g)$,
where $f$ is the conventional $c$-class classifier and 
$g$ is the selection function that reveals the underlying uncertainty of inputs.
Given an input $\x$, selective classifier outputs
\begin{equation}
\label{eq:selective_classifier}
    (f, g)(\x) = \begin{cases}
                \mathrm{Abstain}, & g(\x) > \tau; \\
                f(\x), & \mathrm{otherwise,}
                \end{cases}
\end{equation}
for a given threshold $\tau$ that controls the trade-off.

\subsection{Approach}
Inspired by~\cite{thulasidasan2019dac,ziyindeepgambler},
we adapt our presented approach in Algorithm~\ref{alg:approach} to the selective classification task.
We introduce an extra ($c+1$)-th class (represents \emph{abstention}) during training
and replace selection function $g(\cdot)$ in 
Equation~\eqref{eq:selective_classifier} by $f(\cdot)_{c}$.
In this way, we can train a selective classifier in an end-to-end fashion.
Besides, unlike previous works that provide no explicit signal for learning abstention class,
we use model predictions as a guideline in the design of learning process.
Given  a mini-batch of data pairs $\{(\x_i, \y_i)\}_m$,
model predictions $\p_i$
and its exponential moving average $\bm{t}_i$ for each sample,
we optimize the classifier $f$ by minimizing:
\begin{equation}
\label{eq:abstain_loss}
  \mathcal{L}(f)=-\frac{1}{m}\sum_i [\bm{t}_{i,y_i} \log \p_{i,y_i}+(1-\bm{t}_{i,y_i}) \log \p_{i,c}],
\end{equation}
where $y_i$ is the index of non-zero element in the one hot label vector $\y_i$. The first term measures the cross-entropy loss between prediction and original label $\y_i$, in order to learn a good multi-class classifier.
The second term acts as the selection function, identifies uncertain samples in datasets.
$\bm{t}_{i,y_i}$ dynamically trades-off these two terms:
if $\bm{t}_{i,y_i}$ is very small, the sample is deemed as uncertain and the second term enforces the selective classifier to learn to abstain this sample;
if $\bm{t}_{i,y_i}$ is close to 1, the loss recovers the standard cross entropy minimization
and enforces the selective classifier to make perfect prediction.

\begin{table}[t]
\caption{Selective classification error rate (\%) on CIFAR10 and Dogs vs. Cats datasets
for various coverage rates~(\%). Mean and standard deviation are calculated over 3 trials.
The best entries and those overlap with them are marked bold.}
\label{tab:sel_cls}
\begin{center}
\begin{small}
\begin{tabular}{lcccccc}
\toprule
Dataset & Coverage& Ours & Deep Gamblers & SelectiveNet & SR & MC-dropout \\
\midrule
\multirow{7}{*}{CIFAR10} &  100 & 6.05$\pm$0.20 & 6.12$\pm$0.09 & 6.79$\pm$0.03 & 6.79$\pm$0.03 & 6.79$\pm$0.03  \\
 &  95 & \textbf{3.37$\pm$0.05} & \textbf{3.49$\pm$0.15} & 4.16$\pm$0.09 & 4.55$\pm$0.07 & 4.58$\pm$0.05  \\
 &  90 & \textbf{1.93$\pm$0.09} & 2.19$\pm$0.12 & 2.43$\pm$0.08 & 2.89$\pm$0.03 & 2.92$\pm$0.01  \\
 &  85 & \textbf{1.15$\pm$0.18} & \textbf{1.09$\pm$0.15} & 1.43$\pm$0.08 & 1.78$\pm$0.09 & 1.82$\pm$0.09  \\
 &  80 & \textbf{0.67$\pm$0.10} & \textbf{0.66$\pm$0.11} & 0.86$\pm$0.06 & 1.05$\pm$0.07 & 1.08$\pm$0.05  \\
 &  75 & \textbf{0.44$\pm$0.03} & 0.52$\pm$0.03 & \textbf{0.48$\pm$0.02} & 0.63$\pm$0.04 & 0.66$\pm$0.05  \\
 &  70 & \textbf{0.34$\pm$0.06} & 0.43$\pm$0.07 & \textbf{0.32$\pm$0.01} & 0.42$\pm$0.06 & 0.43$\pm$0.05 \\
\midrule
\multirow{5}{*}{Dogs vs. Cats}  & 100 & 3.01$\pm$0.17 & 2.93$\pm$0.17 & 3.58$\pm$0.04 & 3.58$\pm$0.04 & 3.58$\pm$0.04  \\
 &  95 & \textbf{1.25$\pm$0.05} & \textbf{1.23$\pm$0.12} & 1.62$\pm$0.05 & 1.91$\pm$0.08 & 1.92$\pm$0.06  \\
 &  90 & \textbf{0.59$\pm$0.04} & \textbf{0.59$\pm$0.13} & 0.93$\pm$0.01 & 1.10$\pm$0.08 & 1.10$\pm$0.05  \\
 &  85 & \textbf{0.25$\pm$0.11} & 0.47$\pm$0.10 & 0.56$\pm$0.02 & 0.82$\pm$0.06 & 0.78$\pm$0.06  \\
 &  80 & \textbf{0.15$\pm$0.06} & 0.46$\pm$0.08 & 0.35$\pm$0.09 & 0.68$\pm$0.05 & 0.55$\pm$0.02 \\
\bottomrule
\end{tabular}
\end{small}
\end{center}
\end{table}

\subsection{Experiments}
We conduct the experiments on two datasets: CIFAR10 \cite{krizhevsky2009cifar} and Dogs vs. Cats \cite{catsdogs}.
We compare our method with previous state-of-the-art methods on selective classification, including Deep Gamblers~\cite{ziyindeepgambler}, SelectiveNet~\cite{geifman2019selectivenet}, Softmax Response (SR) and MC-dropout~\cite{geifman2017selective}.
We use the same experimental settings as these works for fair comparison (details are given in Appendix~\ref{sec:setup_selective}).
The results of prior methods are cited from original papers and are summarized in Table~\ref{tab:sel_cls}.
We see that our method achieves up to 50\% relative improvements compared with all other methods under various coverage rates, on all datasets.
Notably, Deep Gamblers also introduces an extra abstention class in their method but without applying model predictions.
The improved performance of our method comes from the use of model predictions in the training process.

\section{Related Works}

\noindent\textbf{Generalization of deep networks}\quad
Previous work~\cite{zhang2016understanding} systematically analyzed the capability of deep networks to overfit random noise. Their results show that traditional wisdom fails to explain the generalization of deep networks.
Another line of works~\cite{opper1995statistical,opper2001learning,advani2017high,spigler2018jamming,belkin2018reconciling,geiger2019jamming,nakkiran2019deep} observed an intriguing double-descent risk curve from the bias-variance trade-off. \cite{belkin2018reconciling,nakkiran2019deep} claimed that this observation challenges the conventional U-shaped risk curve in the textbook. Our work shows that this observation may stem from overfitting of noises; the phenomenon vanishes by a proper design of training process such as self-adaptive training.
To improve the generalization of deep networks, \cite{szegedy2016rethinking,pereyra2017regularizing} proposed label smoothing regularization that uniformly distributes $\epsilon$ of labeling weight to all classes and uses this soft label for training; \cite{zhang2017mixup} introduced mixup augmentation that extends the training distribution by dynamic interpolations between random paired input images and the associated targets during training.
This line of research is similar with ours as both methods use soft labels in the training. However, self-adaptive training is able to recover true labels from noisy labels and is more robust to noises.

\noindent\textbf{Robust learning from corrupted data}\quad
Aside from the approaches that have been discussed in the last paragraph of Section~\ref{sec:contribution}, there have also been many other works on learning from noisy data.
To name a few, \cite{arpit2017closer,li2019gradient} showed that deep neural networks tend to fit clean samples first and overfitting of noise occurs in the later stage of training.
\cite{li2019gradient} further proved that early stopping
can mitigate the issues that are caused by label noises.
\cite{reed2014training,dong2019distillation} incorporated model predictions into training by simple interpolation of labels and model predictions.
We demonstrate that our exponential moving average and sample re-weighting schemes enjoy superior performance.
Other works~\cite{zhang2018gce,wang2019symmetric} proposed alternative loss functions to cross entropy that are robust to label noise. They are orthogonal to ours and are ready to cooperate with our approach as shown in Appendix~\ref{sec:sat_sce}.
Beyond the corrupted data setting, recent works~\cite{furlanello2018born,xie2020self} propose self-training scheme that also uses model predictions as training target. However, they suffers from the heavy cost of multiple iterations of training, which is avoided by our approach.
\section{Conclusion}
In this paper, we study the generalization of deep networks.
We analyze the standard training dynamic using ERM and characterize its intrinsic failure cases under data corruptions.
Our observations motivate us to propose Self-Adaptive Training---a new training algorithm that incorporates model predictions into training process.
We demonstrate that our approach improves the generalization of deep networks under various kinds of corruptions.
Finally, we present two applications of self-adaptive training on classification with label noise and selective classification, where our approach significantly advances the state-of-the-art.


\section*{Broader Impact}
Our work advances robust learning from data under potential corruptions, which is a common feature for real-world, uncurated large-scale datasets due to the error-prone nature of data acquisition.
In contrast to a large existing literature focuses on noisy label setting, our motivation is to provide a generic algorithm that not only is robust to various kinds of noises with varying noise levels, but also incurs no extra computational cost. In practice, these factors are crucial since the exact noise scheme is unknown and the computation budget is indeed very limited.
Built upon the analysis on the intrinsic failure patterns of ERM under data corruptions, we introduce an elegant way to incorporate model predictions into training process to improve the generalization of deep networks under noisy data.
By correcting outliers and calibrating the training process, our approach is ready for real-world application of deep learning. It can serve as a basic building block of large-scale AI system that generalizes well to a wide range of visual tasks.

While we have empirically evaluated our approach on data under both random and adversarial noises, most of our studies focus on artificial data corruptions (except the analysis on ImageNet which contains annotation error by itself), which may not represent natural noises in practice. However, the presented methodology that properly incorporates model predictions into training process sheds light on understanding and improving the generalization of deep networks under data corruptions 
in the future study.

\begin{ack}
This work was supported in part by the National Nature Science Foundation of China under Grant 62071013 and 61671027. Hongyang Zhang was supported in part by the Defense Advanced Research Projects Agency under cooperative agreement HR00112020003.
\end{ack}

{\small
\bibliography{paper_final.bbl}
\bibliographystyle{ieee}
}

\onecolumn
\newpage
\appendix
\addappheadtotoc

\begin{figure}[t]
    \centering
    \begin{subfigure}[t]{\textwidth}
        \centering
        \includegraphics[width=\textwidth]{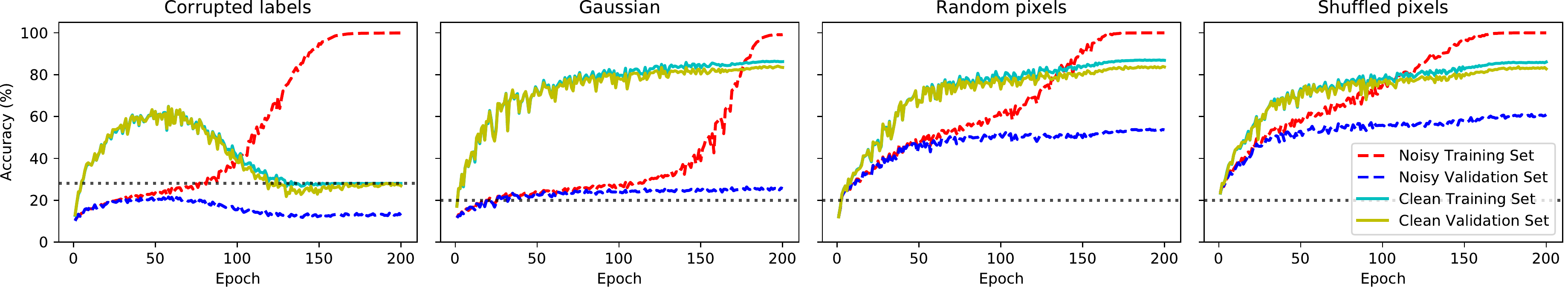}
        \caption{Accuracy curves of model trained using ERM.}
        \label{fig:ce_acc_curve_r08}
    \end{subfigure}
    \vskip 0.075in
    \begin{subfigure}[t]{\textwidth}
        \centering
        \includegraphics[width=\textwidth]{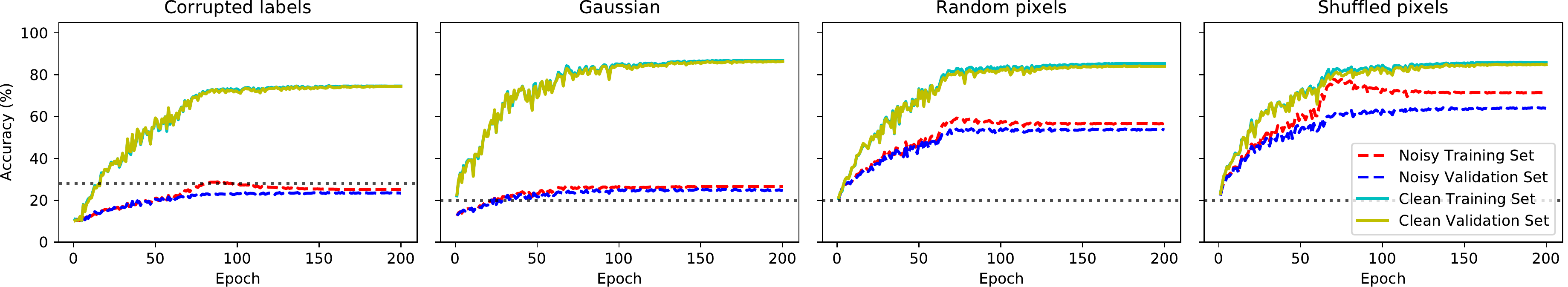}
        \caption{Accuracy curves of model trained using our method.}
        \label{fig:wsc_acc_curve_r08}
    \end{subfigure}
    \caption{
    Accuracy curves of model trained on noisy CIFAR10 training set with 80\% noise rate. 
    The horizontal dotted line displays the percentage of clean data in the training sets.
    It shows that our observations in Section~\ref{sec:approach} hold true even when extreme label noise injected.
    }
    \label{fig:acc_curve_r08}
\end{figure}

\section{Experimental Setups}
\subsection{Double descent phenomenon}
\label{sec:setup_dd}
Following previous work~\cite{nakkiran2019deep}, 
we optimize all models using Adam~\cite{kingma2014adam} optimizer
with fixed learning rate of 0.0001, batch size of 128, common data augmentation, weight decay of 0 for 4,000 epochs.
For our approach, we use the hyper-parameters $\mathrm{E}_s=40, \alpha=0.9$ for standard ResNet-18 (width of 64) and dynamically adjust them for other models according to the relation of model capacity $r=\frac{64}{\mathrm{width}}$ as:
\begin{equation}
\label{eq:dd_params}
    \mathrm{E}_s = 40\times r;\quad \alpha = 0.9^{\frac{1}{r}}.
\end{equation}

\subsection{Adversarial training}
\label{sec:setup_adv}
\cite{szegedy2013intriguing} reported that imperceptible small perturbations around input data (i.e., adversarial examples) can cause ERM trained deep neural networks to make arbitrary predictions.
Since then, a large literature devoted to improving the adversarial robustness of deep neural networks.
Among them, adversarial training algorithm TRADES \cite{zhang2019theoretically} achieves state-of-the-art performance.
TRADES decomposed robust error (w.r.t adversarial examples) to sum of natural error and boundary error, and proposed to minimize:
\begin{equation}
\label{eq:trades_appendix}
\mathbb{E}_{\x,\y}\Bigg\{ \mathrm{CE}(\p(\x), \y) + \max_{\|\widetilde{\x}-\x\|_\infty\le\epsilon}\mathrm{KL}(\p(\x), \p(\widetilde{\x}))/\lambda\Bigg\},
\end{equation}
where $\p(\cdot)$ is the model prediction, $\epsilon$ is the maximal allowed perturbation, CE stands for cross entropy, KL stands for Kullback–Leibler divergence.
The first term corresponds to ERM that maximizes the natural accuracy; the second term pushes the decision boundary away from data points to improve adversarial robustness; the hyper-parameter $1/\lambda$ controls the trade-off between natural accuracy and adversarial robustness.
We evaluate self-adaptive training on this task by replacing the first term of Equation~\eqref{eq:trades_appendix} with our approach.

Our experiments are based on the official open-sourced implementation\footnote{\url{https://github.com/yaodongyu/TRADES}} of TRADES~\cite{zhang2019theoretically}.
Concretely, we conduct experiments on CIFAR10 dataset \cite{krizhevsky2009cifar} and use WRN-34-10~\cite{zagoruyko2016wide} as base classifier.
For training, we use initial learning rate of 0.1, batch size of 128, 100 training epochs. The learning rate is decayed at 75-th, 90-th epoch by a factor of 0.1.
The adversarial example $\widetilde{\x}_i$ is generated dynamically during training by projected gradient descent (PGD) attack \cite{madry2017towards} with maximal $\ell_{\infty}$ perturbation $\epsilon$ of 0.031, perturbation step size of 0.007, number of perturbation steps of 10.
The hyper-parameter $1/\lambda$ of TRADES is set to 6 as suggested by original paper, $\mathrm{E}_s, \alpha$ of our approach is set to 70, 0.9, respectively.
For evaluation, we report robust accuracy $\frac{1}{n}\sum_i \mathbbm{1}\{ \argmax~p(\widetilde{\x}_i)=\argmax~\y_i \}$, where adversarial example $\widetilde{\x}$ is generated by white box $\ell_{\infty}$ untargeted PGD attack with $\epsilon$ of 0.031, perturbation step size of 0.007, number of perturbation steps of 20.

\subsection{ImageNet}
\label{sec:setup_imagenet}
We use ResNet-50~\cite{he2016deep} as base classifier.
Following original paper~\cite{he2016deep} and~\cite{loshchilov2016sgdr,goyal2017accurate}, we use SGD to optimize the networks with batch size of 768, base learning rate of 0.3, momentum of 0.9, weight decay of 0.0005 and total training epoch of 95.
The learning rate is linearly increased from 0.0003 to 0.3 in first 5 epochs (i.e., warmup), and then decayed using cosine annealing schedule~\cite{loshchilov2016sgdr} to 0.
Following common practice, we use random resizing, cropping and
flipping augmentation during training.
The hyper-parameters of our approach are set to $\mathrm{E}_s = 50$ and $\alpha=0.99$ under standard setup, and are set to $\mathrm{E}_s = 60$ and $\alpha = 0.95$ under 40\% label noise setting. The experiments are conducted on PyTorch~\cite{paszke2019pytorch} with distributed training and mixed precision training\footnote{\url{https://github.com/NVIDIA/apex}} for acceleration.

\subsection{Selective classification}
\label{sec:setup_selective}
The experiments are base on official open-sourced implementation\footnote{\url{https://github.com/Z-T-WANG/NIPS2019DeepGamblers}} of Deep Gamblers to ensure fair comparison.
We use the VGG-16 network~\cite{simonyan2014very} with batch normalization~\cite{ioffe2015batch} and dropout~\cite{srivastava2014dropout} as base classifier in all experiments.
The network is optimized using SGD with initial learning rate of 0.1, momentum of 0.9, weight decay of 0.0005, batch size of 128, total training epoch of 300.
The learning rate is decayed by 0.5 in every 25 epochs.
For our method, we set the hyper-parameters $\mathrm{E}_s=0, \alpha=0.99$.

\begin{figure}[t]
    \centering
    \includegraphics[width=.65\textwidth]{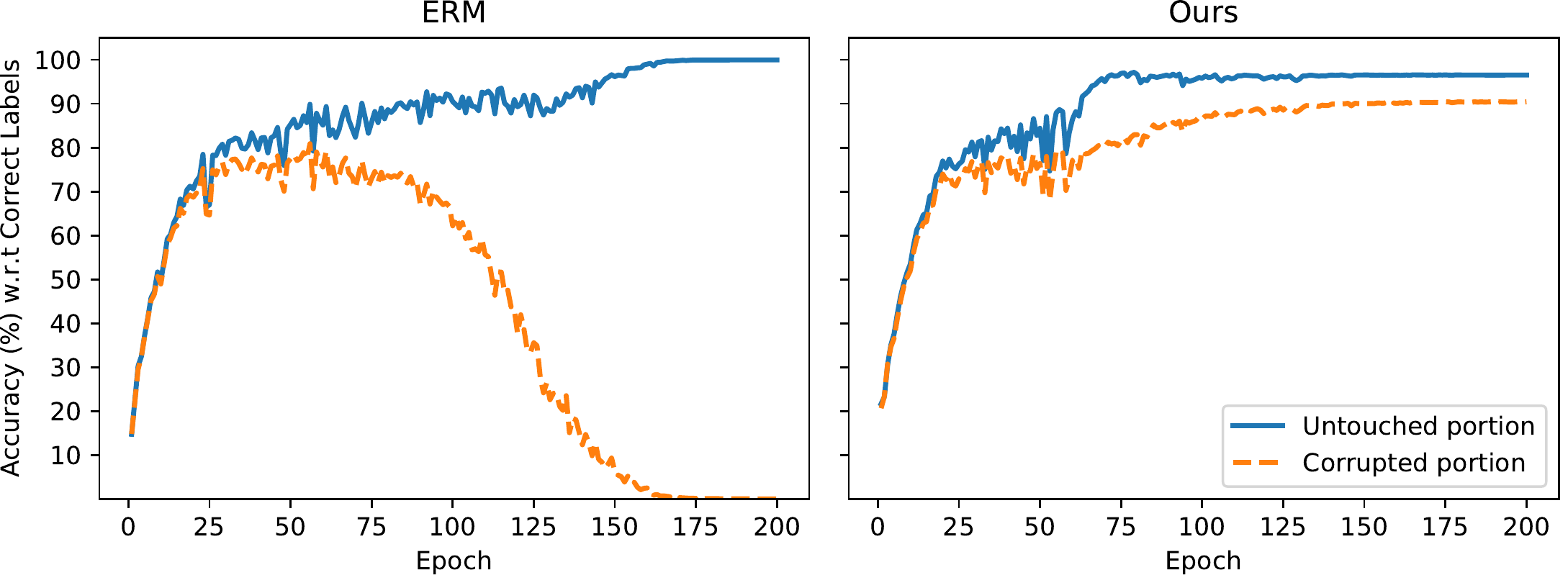}
    \caption{
    Accuracy curves on different portions of the CIFAR10 training set (with 40\% label noise) w.r.t. correct labels. We split the training set into two portions:
    1)~\emph{Untouched portion}, i.e., the elements in the training set which were left untouched; 
    2)~\emph{Corrupted portion}, i.e., the elements in the training set which were indeed randomized. It shows that ERM fits correct labels in the first few epochs and then eventually overfits the corrupted labels. In contrast, self-adaptive training calibrates the training process and consistently fits the correct labels.
    }
    \label{fig:split_acc_curve_r04}
\end{figure}

\section{Additional Experimental Results \& Discussions}

\subsection{ERM may suffer from overfitting of noise}

In~\cite{zhang2016understanding}, the authors showed that
the model trained by standard ERM can easily fit randomized data. However, they only analyzed the generalization errors in the presence of corrupted labels. In this paper, we report the whole training process and also consider the performance on clean sets (i.e., the original uncorrupted data). Figure~\ref{fig:ce_acc_curve} shows the four accuracy curves (on clean and noisy training, validation set, respectively) for each model that is trained on one of four corrupted training data. Note that the models can only have access to the noisy training sets (i.e., the red curve) and the other three curves are shown only for the illustration purpose. We conclude with two principal observations from the figures:
(1) The accuracy on noisy training and validation sets is close at beginning and the gap is monotonously increasing w.r.t. epoch. The generalization errors (i.e., the gap between the accuracy on noisy training and validation sets) are large at the end of training.
(2) The accuracy on clean training and validation set is consistently higher than the percentage of clean data in the noisy training set. This occurs around the epochs between underfitting and overfitting.

Our first observation poses concerns on the overfitting issue of ERM training dynamic which has also been reported by~\cite{li2019gradient}. However, the work of~\cite{li2019gradient} only considered the case of corrupted labels and proposed using early-stop mechanism to improve the performance on clean data. On the other hand, our analysis of the broader corruption schemes shows that the early stopping might be sub-optimal and may hurt the performance under other types of corruptions (see the last three columns in Figure~\ref{fig:ce_acc_curve}).

The second observation implies that, perhaps surprisingly, model predictions by ERM can capture and amplify useful signals in the noisy training set, although the training dataset is heavily corrupted. While this was also reported in~\cite{zhang2016understanding,rolnick2017deep,guan2018said,li2019gradient} for the case of corrupted labels, we show that similar phenomenon occurs under other kinds of corruptions more generally. This observation sheds light on our approach, which incorporates model predictions into training procedure.

\subsection{Improved generalization of self-adaptive training on random noise}
\noindent\textbf{Training accuracy w.r.t. correct labels on different portions of data}\quad
For more intuitive demonstration, we split the CIFAR10 training set (with 40\% label noise) into two portions:
1)~\emph{Untouched portion}, i.e., the elements in the training set which were left untouched;
2)~\emph{Corrupted portion}, i.e., the elements in the training set which were indeed randomized.
The accuracy curves on these two portions w.r.t correct training labels is shown in Figure~\ref{fig:split_acc_curve_r04}.
We can observe that the accuracy of ERM on the corrupted portion first increases in the first few epochs and then eventually decreases to 0. In contrast, self-adaptive training calibrates the training process and consistently fits the correct labels.

\noindent\textbf{Study on extreme noise}\quad
We further rerun the same experiments as in Figure~\ref{fig:acc_curve} of main text by injecting extreme noise (i.e., noise rate of 80\%) into CIFAR10 dataset. We report the corresponding accuracy curves in Figure~\ref{fig:acc_curve_r08}, which shows that our approach significantly improves the generalization over ERM even when random noise dominates training data. This again justify our observations in Section~\ref{sec:approach} of the main body.

\begin{figure}[t]
\centering
\includegraphics[width=\textwidth]{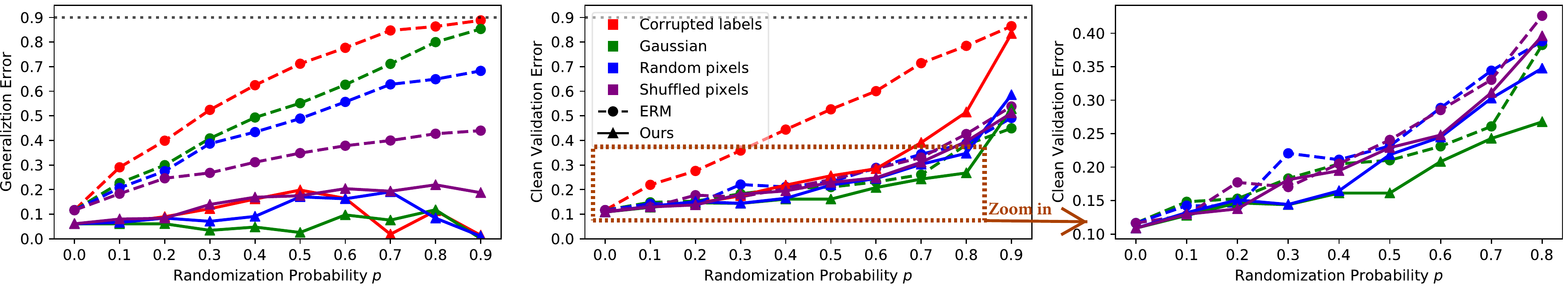}
\caption{
Generalization error and clean validation error under four random noises (represented by different colors) for ERM (the dashed curves) and our approach (the solid curves) on CIFAR10 when data augmentation is turned off. We zoom-in the dashed rectangle region and display it in the third column for clear demonstration.
}
\label{fig:gen_clean_errs_wo_aug}
\end{figure}

\begin{figure}[t]
    \centering
    \includegraphics[width=.45\textwidth]{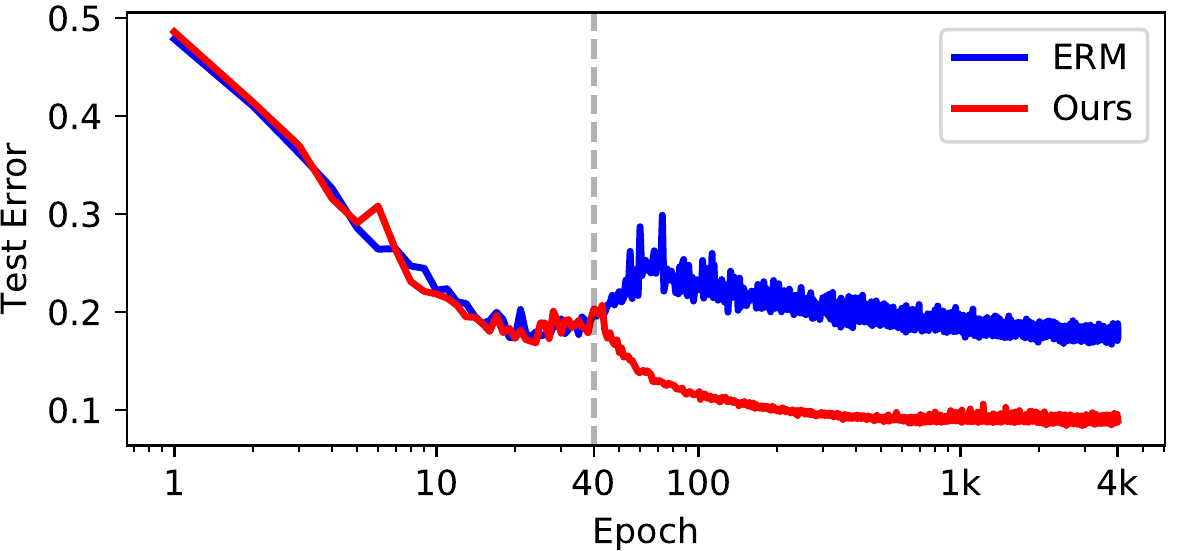}
    \caption{Self-adaptive training \emph{vs.} ERM on the error-epoch curve.
    We train the standard ResNet-18 networks (i.e., width of 64) on the CIFAR10 dataset with 15\% randomly-corrupted labels and report the test errors on the clean data.
    The dashed vertical line represents the initial epoch $\mathrm{E}_s$ of our approach.
    It shows that self-adaptive training has significantly diminished epoch-wise double-descent phenomenon.
    }
    \label{fig:epochwise_dd}
\end{figure}

\begin{table}[t]
\caption{Parameters sensitivity to different datasets and noise rates.}
\label{tab:ablation_alpha_es_supp}
\begin{center}
\begin{small}
\setlength{\tabcolsep}{1.5mm}{
\begin{tabular}{lcccccc}
\toprule
& \multicolumn{3}{c}{CIFAR10 (80\% Noise)} & \multicolumn{3}{c}{CIFAR100 (40\% Noise)} \\
$\alpha$ & 0.8 & 0.9 & 0.95 & 0.8 & 0.9 & 0.95 \\
\midrule
Fix $\mathrm{E}_s$=60 & 75.60 & \textbf{78.58} & 75.44 & 70.36 & \textbf{71.38} & 68.57 \\
\midrule\midrule
$\mathrm{E}_s$ & 40 & 60 & 80 & 40 & 60 & 80 \\
\midrule
Fix $\alpha$=0.9 & 68.27 & 78.58 & \textbf{78.65} & 70.30 & \textbf{71.38} & 67.32 \\
\bottomrule
\end{tabular}
}
\end{small}
\end{center}
\end{table}

\begin{table}[t]
\caption{Test Accuracy (\%) on CIFAR datasets with various levels of uniform label noise injected to training set.
We show that considerable gains can be obtained when combined with SCE loss.}
\label{tab:noisy_cls_ours_sce}
\begin{center}
\begin{small}
\begin{tabular}{lcccccccc}
\toprule
 & \multicolumn{4}{c}{CIFAR10} & \multicolumn{4}{c}{CIFAR100} \\
\multirow{2}{*}{Method} & \multicolumn{4}{c}{Label Noise Rate} & \multicolumn{4}{c}{Label Noise Rate} \\
        & 0.2 & 0.4 & 0.6 & 0.8 & 0.2 & 0.4 & 0.6 & 0.8   \\
\midrule
SCE~\cite{wang2019symmetric} & 90.15 & 86.74 & 80.80 & 46.28 & 71.26 & 66.41 & 57.43 & 26.41\\
Ours    & 94.14 & 92.64 & 89.23 & 78.58 & 75.77 & 71.38 & 62.69 & 38.72 \\
Ours + SCE & \textbf{94.39} & \textbf{93.29} & \textbf{89.83} & \textbf{79.13} & \textbf{76.57} & \textbf{72.16} & \textbf{64.12} & \textbf{39.61} \\
\bottomrule
\end{tabular}
\end{small}
\end{center}
\end{table}

\noindent\textbf{Effect of data augmentation}\quad
All our previous studies are performed with common data augmentation (i.e., random cropping and flipping). Here, we further report the effect of data augmentation. We adjust introduced hyper-parameters as $\mathrm{E}_s=25$, $\alpha=0.7$ due to severer overfitting when data augmentation is absent. The Figure~\ref{fig:gen_clean_errs_wo_aug} shows the corresponding generalization errors and clean validation errors. We observe that, for both ERM and our approach, the errors clearly increase when data augmentation is absent (compared with those in Figure~\ref{fig:gen_clean_errs}). However, the gain is limited and the generalization errors can still be very large, with or without data augmentation for standard ERM. Directly replacing the standard training procedure with our approach can bring bigger gains in terms of generalization regardless of data augmentation. This suggests that data augmentation can help but is not of essence to improve generalization of deep neural networks, which is consistent with the observation in~\cite{zhang2016understanding}.

\subsection{Epoch-wise double descent phenomenon}
Prior work~\cite{nakkiran2019deep} reported that, for sufficient large model,
test error-training epoch curve also exhibits double-descent phenomenon,
which they termed \emph{epoch-wise double descent}.
In Figure~\ref{fig:epochwise_dd}, we reproduce the epoch-wise double descent phenomenon on ERM
and inspect self-adaptive training.
We observe that our approach (the red curve) exhibits slight double-descent
due to overfitting starts before initial $\mathrm{E}_s$ epochs.
As the training targets being updated (i.e., after $\mathrm{E}_s$ = 40 training epochs),
the red curve undergoes monotonous decrease.
This observation again indicates that double-descent phenomenon 
may stem from overfitting of noise and can be avoided by our algorithm.

\subsection{Cooperation with symmetric cross entropy}
\label{sec:sat_sce}
Prior work~\cite{wang2019symmetric} showed that Symmetric Cross Entropy (SCE) loss is robust to underlying label noise in training data. Formally, given training target $\bm{t}_i$ and model prediction $\p_i$, SCE loss is defined as:
\begin{equation}
    \mathcal{L}_{sce} = -w_1 \sum_j \bm{t}_{i,j}~\log~\p_{i,j} - w_2 \sum_j \bm{p}_{i,j}~\log~\bm{t}_{i,j},
\end{equation}
where the first term is the standard cross entropy loss and the second term is the reversed version. In this section, we show that self-adaptive training can cooperate with this noise-robust loss and enjoy further performance boost without extra cost.

\noindent\textbf{Setup}\quad
The most experiments settings are kept the same as Section~\ref{sec:exp_label_noise}.
For the introduced hyper-parameters $w_1, w_2$ of SCE loss, we directly set them to 1, 0.1, respectively, in all our experiments for simplicity.

\noindent\textbf{Results}\quad
We summarize the results in Table~\ref{tab:noisy_cls_ours_sce}.
We cam see that, although self-adaptive training already achieves very strong performance, considerable gains can be obtained when equipped with SCE loss. Concretely, the improvement is as large as 1.5\% when label noise of 60\% injected to CIFAR100 training set. It also indicates that our approach is flexible and is ready to cooperate with alternative loss functions.

\begin{table}[t]
\caption{Average Accuracy (\%) on CIFAR10 test set and out-of-distribution dataset CIFAR10-C at various corruption levels.}
\label{tab:cifar10c}
\begin{center}
\begin{small}
\begin{tabular}{lcccccc}
\toprule
\multirow{2}{*}{Method} & \multirow{2}{*}{CIFAR10} & \multicolumn{5}{c}{Corruption Level@CIFAR10-C} \\
\cmidrule{3-7}
 & & 1 & 2 & 3 & 4 & 5 \\
\midrule
ERM     & 95.32 & 88.44 & 83.22 & 77.26 & 70.40 & 58.91 \\
Ours    & \textbf{95.80} & \textbf{89.41} & \textbf{84.53} & \textbf{78.83} & \textbf{71.90} & \textbf{60.77} \\
\bottomrule
\end{tabular}
\end{small}
\end{center}
\end{table}

\subsection{Out-of-distribution generalization}
In this section, we consider out-of-distribution (OOD) generalization, where the models are evaluated on unseen test distributions outside the training distribution. 

\noindent\textbf{Setup}\quad
To evaluate the OOD generalization performance, we use CIFAR10-C benchmark~\cite{hendrycks2018benchmarking} that constructed by applying 15 types of corruption to the original CIFAR10 test set at 5 levels of severity. The performance is measure by average accuracy over 15 types of corruption. We mainly follow the training details in Section~\ref{sec:exp_label_noise} and adjust $\alpha=0.95, \mathrm{E}_s=80$.

\noindent\textbf{Results}\quad
We summarize the results in Table~\ref{tab:cifar10c}. Regardless the presence of corruption and corruption levels, our method consistently outperforms ERM by a considerable margin, which becomes large when the corruption is more severe. The experiment indicates that self-adaptive training may provides implicit regularization for OOD generalization.

\subsection{Cost of maintaining probability vectors}
Take the large-scale ImageNet dataset~\cite{deng2009imagenet} as an example. The ImageNet consists of about 1.2 million images categorized to 1000 classes. The storage of such vectors in single precision format for the entire dataset requires $1.2 \times 10^6 \times 1000 \times 32$ bit $\approx 4.47$GB, which is acceptable since modern GPUs usually have no less than 11GB memory. Moreover, the vectors can be stored on CPU memory or even disk and loaded along with the images to further reduce the cost.

\end{document}